\documentclass{svproc}
\usepackage{subfiles}
\usepackage{url}

\usepackage{graphicx}
\usepackage{amsmath}
\usepackage[table]{xcolor}
\usepackage{amssymb}
\usepackage{mathtools}
\usepackage{algorithm}
\usepackage[noend]{algpseudocode}
\usepackage{algorithmicx}

\usepackage{pgfplots}
\usepackage{wrapfig}
\usepackage{hyperref}
\hypersetup{
	colorlinks,
	linkcolor={red!65!black},
	citecolor={green!65!black},
	urlcolor={blue!65!black}
}

\usetikzlibrary{matrix}
\usetikzlibrary{backgrounds}
\usetikzlibrary{calc}
\usetikzlibrary{shapes.misc}

\usepackage{subcaption}
\usepackage{scalefnt}

\newcommand{\bs}[1]{\boldsymbol{#1}}
\newcommand{\bno}[2]{\bs{#1}_{#2}}
\newcommand{\btx}[2]{\bs{#1}_{\text{#2}}}
\newcommand{\inv}[1]{#1^\dag}
\newcommand{\qno}[1]{\bno{q}{#1}}
\newcommand{\qtx}[1]{\btx{q}{#1}}
\newcommand{\xno}[1]{\bno{x}{#1}}
\newcommand{\xtx}[1]{\btx{x}{#1}}
\newcommand{\vno}[1]{\bno{v}{#1}}
\newcommand{\lano}[1]{\bno{\lambda}{#1}}
\newcommand{\hT}{^\text{T}}

\newcommand{\wno}[1]{\bno{\omega}{#1}}

\newcommand\ddfrac[2]{\frac{\displaystyle #1}{\displaystyle #2}}
\newcommand{\derivb}[2]{\ddfrac{\partial#1}{\partial#2}}

\newcommand{\Lmat}[1]{L(#1)}
\newcommand{\LTmat}[1]{\Lmat{#1}\hT}
\newcommand{\Rmat}[1]{R(#1)}
\newcommand{\RTmat}[1]{\Rmat{#1}\hT}

\newcommand{\dt}{\Delta t}
\newcommand{\itind}[1]{^{(#1)}}

\newcommand{\cellcr}{\cellcolor{red!15}}

\newenvironment{valgorithm}{\vspace*{-2mm}\begin{algorithm}}{\end{algorithm}\vspace*{-2mm}}

\newenvironment{vfigure}{\begin{figure}}{\end{figure}\vspace*{-3mm}}

\begin{document}
	
\mainmatter              
\title{Linear-Time Variational Integrators in Maximal Coordinates}
\titlerunning{Linear-Time Variational Integrators in Maximal Coordinates}  
%
\author{Jan Br\"udigam \and Zachary Manchester}
\authorrunning{Jan Br\"udigam and Zachary Manchester} 
%
\tocauthor{Jan Br\"udigam and Zachary Manchester}
\institute{Stanford University, Stanford CA 94305, USA,\\
	\email{\{bruedigam, zacm\}@stanford.edu}
	}

\maketitle              

\begin{abstract}
	Most dynamic simulation tools parameterize the configuration of multi-body robotic systems using minimal coordinates, also called generalized or joint coordinates. However, maximal-coordinate approaches have several advantages over minimal-coordinate parameterizations, including native handling of closed kinematic loops and nonholonomic constraints. This paper describes a linear-time variational integrator that is formulated in maximal coordinates. Due to its variational formulation, the algorithm does not suffer from constraint drift and has favorable energy and momentum conservation properties. A sparse matrix factorization technique allows the dynamics of a loop-free articulated mechanism with $n$ links to be computed in $O(n)$ (linear) time. Additional constraints that introduce loops can also be handled by the algorithm without incurring much computational overhead. Experimental results show that our approach offers speed competitive with state-of-the-art minimal-coordinate algorithms while outperforming them in several scenarios, especially when dealing with closed loops and configuration singularities.
\end{abstract}
%
\section{Introduction}
In many fields, the predominant method to describe rigid body dynamics is with \textit{minimal coordinates} (also called joint or generalized coordinates). In this approach, each degree of freedom of a system is represented by a single coordinate describing, for example, an angle or a displacement. The main reasons for this choice of coordinates are computational performance---since only the minimal required set of variables are tracked---the absence of explicit constraints to enforce joint behavior, and the prevention of constraint drift, i.e. links of a mechanism drifting apart due to numerical integration errors \cite{baumgarte_stabilization_1972}.

In contrast, a different way of describing configurations of rigid bodies is with \textit{maximal coordinates}, in which all six degrees of freedom of each body are modeled and the reduction of degrees of freedom of a system is achieved by imposing constraints via Lagrange multipliers. While linear-time computational performance for loop-free structures has been proved for both minimal and maximal coordinates more than 20 years ago \cite{Featherstone08,baraff_linear-time_1996}, constraint drift when using maximal coordinates is still an issue and requires stabilization schemes \cite{baumgarte_stabilization_1972,macklin_unified_2014}. 

A separate and more recent area of research regarding rigid body dynamics is concerned with \textit{variational integrators} \cite{marsden_discrete_2001}. These integrators have a number of advantages over classical Runge-Kutta methods when numerically integrating the differential equations used to describe mechanical systems. Instead of deriving the equations of motion of a dynamical system in continuous time and then discretizing them, variational approaches discretize the derivation of the equations of motion itself, and thereby retain many of the properties of the real system such as energy and momentum conservation \cite{marsden_discrete_2001,hairer_geometric_2006}.

In the last few years, algorithms for calculating rigid body dynamics in linear time with variational integrators in minimal coordinates have been presented \cite{lee_linear-time_2016,fan_efficient_2018}. While these algorithms show promising results, mechanical structures with closed kinematic loops or nonholonomic (i.e. velocity dependent) constraints have not been explicitly treated. Additionally, closed-loop structures and nonholonomic settings require explicit constraints, which diminish the advantages of minimal coordinates since these constraints can no longer be eliminated in advance and have to be handled similarly to the maximal-coordinate approach.

To address the accurate handling of constrained mechanical systems, this paper presents:
\begin{itemize}
	\item A variational integrator for the equations of motion of rigid bodies in maximal coordinates that guarantees constraint satisfaction.
	\item An algorithm to calculate the dynamics of loop-free mechanical structures with this integrator in linear time.
	\item The capability to robustly extend this algorithm by loop-closure constraints with limited increase in computation time.
\end{itemize}

This paper is structured as follows: After introducing our notation and some background in Sec. \ref{sec:background}, we derive a variational integrator in maximal coordinates for the equations of motion of rigid bodies in Sec.  \ref{sec:variational}. Subsequently, in Sec.  \ref{sec:solver} we state the algorithms to integrate the equations of motion in linear-time and show experimental results in Sec.  \ref{sec:results}.

\section{Background} \label{sec:background}
We will give a brief review of quaternions (see \cite{sola_quaternion_2017} for details), and provide an example of two rigid bodies in maximal coordinates and how their connection with a joint can be described by a constraint equation.

\subsection{Quaternions}
Quaternions are a natural choice to represent the orientation of a rigid body as they are globally non-singular as opposed to three-parameter representations, but still have a concise set of only four parameters, unlike, for example, rotation matrices with nine parameters.

Quaternions have four components, commonly written as a stacked vector,
\begin{equation}\label{eqn:quat}
\bs{q}
=
\begin{bmatrix}
q_w \\ q_{v_1} \\ q_{v_2} \\ q_{v_3}
\end{bmatrix}
= 
\begin{bmatrix}
q_w \\ \bs{q}_v
\end{bmatrix},
\end{equation}
where $q_w$ and $\bs{q}_v$ are called the scalar and vector components, respectively. In this paper we will only be using unit quaternions, i.e. $\bs{q}^T\bs{q} = 1$, to represent orientations, and we are using the Hamilton convention with a local-to-global rotation action. In this convention, a quaternion $\bs{q}$ maps vectors from the local to the global frame, whereas its inverse maps from the global to the local frame.

Notation \eqref{eqn:quat} allows for a simple formulation of the basic operations \textit{inverse} and \textit{multiplication}:
\begin{alignat}{2}
&\text{Inverse: } ~~&&\inv{\bs{q}} = 
\begin{bmatrix} 
q_w \\ 
-\qno{v} 
\end{bmatrix} \\
&\text{Multiplication: } ~~\bs{q} \otimes &&\bs{p} = 
\begin{bmatrix}
q_{w} p_{w} - \qno{v}\hT \bs{p}_v 
\\q_{w}\bs{p}_v + p_{w}\qno{v} + \qno{v} \times \bs{p}_v
\end{bmatrix}
\end{alignat}

The $\times$ operator indicates the standard cross product of two vectors. Three other common operations include expanding a vector $\bs{x} \in \mathbb{R}^3$ into a quaternion, retrieving the vector part from a quaternion, and constructing a skew-symmetric matrix from a vector $\bs{x} \in \mathbb{R}^3$ to form the cross product as a matrix-vector product:
\begin{alignat}{2}
&\text{Expand vector: } ~~&&\hat{\bs{x}} = 
\begin{bmatrix} 
0 \\ 
\bs{x} 
\end{bmatrix}\\
&\text{Retrieve vector: } ~~&&\bs{q}^{\text{v}} = \qno{v}
\\
&\text{Skew-symmetric matrix: } ~~&&\bs{x}^{\times} = \begin{bmatrix}0 & -x_3 & x_2\\ x_3 & 0 & -x_1\\ -x_2 & x_1 & 0\end{bmatrix} \rightarrow \xno{1}\times\xno{2} = \xno{1}^{\times}\xno{2}
\end{alignat}

To simplify calculations with quaternions, we introduce the following  four matrices with the identity matrix $I_3 \in \mathbb{R}^{3\times3}$:
\begin{alignat}{2}
T &= \begin{bmatrix}1 &~~ \bs{0}\hT\\\bs{0} &~~ -I_3\end{bmatrix} &&~~ \in \mathbb{R}^{4\times4}\\
\Lmat{\bs{q}} &= \begin{bmatrix}q_w &~~ -\qno{v}\hT\\\qno{v} &~~ q_w I_3 + \qno{v}^{\times} \end{bmatrix} &&~~ \in \mathbb{R}^{4\times4}\\
\Rmat{\bs{q}} &= \begin{bmatrix}q_w &~~ -\qno{v}\hT\\\qno{v} &~~ q_w I_3 - \qno{v}^{\times} \end{bmatrix} &&~~ \in \mathbb{R}^{4\times4}\\
V &= \begin{bmatrix}\bs{0} &~~ I_3\end{bmatrix} &&~~ \in \mathbb{R}^{3\times4}
\end{alignat}
These matrices allow us to perform all required quaternion operations as matrix-vector products for which the standard rules of linear algebra hold:
\begin{align}
\bs{q}_1 \otimes \bs{q}_2 &= L(\bs{q}_1)\bs{q}_2 = R(\bs{q}_2)\bs{q}_1\\
\inv{\bs{q}} &= T\bs{q}\\
\bs{q}^{\text{v}} &= V\bs{q}\\
\hat{\bs{x}} &= V\hT\bs{x}
\end{align}

Now, we can write the rotation of a vector $\bs{x}$ as a matrix-vector product,
\begin{equation}
\left(\bs{q}\otimes\hat{\bs{x}}\otimes\inv{\bs{q}}\right)^{\text{v}} = V \RTmat{\bs{q}} \Lmat{\bs{q}} V\hT \bs{x} = V \Lmat{\bs{q}} \RTmat{\bs{q}} V\hT \bs{x},
\end{equation}
and do the same for the angular velocity:
\begin{equation}\label{eqn:angvel}
\bs{\omega} = 2\left(\inv{\bs{q}}\otimes\dot{\bs{q}}\right)^{\text{v}} = 2 V \LTmat{\bs{q}} \dot{\bs{q}}.
\end{equation}

\subsection{Articulated Rigid Body}

\begin{wrapfigure}{r}{0.5\textwidth}
	\vspace*{-8mm}
	\centering
	\begin{tikzpicture}
	\draw[fill=black!5,rounded corners=6pt,rotate=20]
	(0,0.0) rectangle ++(2,-0.4);
	\draw[fill=black!5,rounded corners=6pt,rotate=-30]
	(1.12,1.45) rectangle ++(2,-0.4);
	
	\draw[fill=black] (1.76,0.43) circle (0.065);
	
	\coordinate (l1) at (0.85,0.1);
	\draw[->,draw=blue!80,thick] (l1) -- ($(l1)+(-0.16,0.4)$);
	\draw[->,draw=blue!80,thick] (l1) -- ($(l1)+(0.4,0.16)$);
	\draw[->,draw=blue!80,thick] (l1) -- ($(l1)+(-0.15,-0.3)$);
	
	\coordinate (l2) at (2.47,-0.02);
	\draw[->,draw=blue!80,thick] (l2) -- ($(l2)+(0.23,0.36)$);
	\draw[->,draw=blue!80,thick] (l2) -- ($(l2)+(0.36,-0.23)$);
	\draw[->,draw=blue!80,thick] (l2) -- ($(l2)+(-0.15,-0.3)$);
	
	\draw[->,draw=black,thick] (l1) -- (1.72,0.41);
	\draw[->,draw=black,thick] (l2) -- (1.8,0.41);
	
	\coordinate (w) at (1.5,-1.2);
	\draw[->,draw=blue!80,thick] (w) -- ($(w)+(0.0,0.42)$);
	\draw[->,draw=blue!80,thick] (w) -- ($(w)+(0.42,0.0)$);
	\draw[->,draw=blue!80,thick] (w) -- ($(w)+(-0.15,-0.3)$);
	
	\node at (1.3,0.7) {$\btx{p}{a}$};
	\node at (2.3,0.6) {$\btx{p}{b}$};
	
	\node at (-0.2,0.5) {Link a};
	\node at (3.7,0.2) {Link b};
	
	\draw (1.3,0.55) -- (1.42,0.29);
	\draw (2.3,0.5) -- (2.1,0.25);
	
	\node at ($(w)+(1.0,-0.3)$) {Global frame};
	
	\draw ($(w)+(-0.07,0.07)$) to[out=150,in=-80 ] ($(l1)+(0.05,-0.05)$);
	\draw ($(w)+(0.07,0.07)$) to[out=50,in=180 ] ($(l2)+(-0.07,-0.02)$);
	
	\node at (0.5,-0.8) {$\xtx{a}$, $\qtx{a}$};
	\node at (2.5,-0.7) {$\xtx{b}$, $\qtx{b}$};
	\end{tikzpicture}
	\caption{Two links connected by a joint.}
	\vspace*{-4mm}
	\label{fig:twolinks}
\end{wrapfigure}
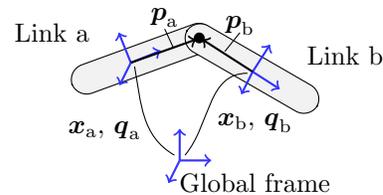
A single rigid body has a position $\bs{x}$, velocity $\bs{v}$, and mass $m$, as well as an orientation $\bs{q}$, angular velocity $\bs{\omega}$, and a moment of inertia matrix $J$. All quantities refer to the center of mass of a body.

For an articulated mechanism consisting of multiple rigid bodies, we can formulate constraints $\bs{g}$ to represent, for instance, joints. As an example, we give the constraints for the ball-and-socket joint connecting the links $\text{a}$ and $\text{b}$ in Fig. \ref{fig:twolinks} (see \cite{siciliano_springer_2016} for a complete list of common joint types):
\begin{equation}
\bs{g} = \bs{x}_{\text{a}} + V \RTmat{\bs{q}_{\text{a}}} \Lmat{\bs{q}_{\text{a}}} V\hT \bs{p}_{\text{a}} ~~ - ~~ \left(\bs{x}_{\text{b}} + V \RTmat{\bs{q}_{\text{b}}} \Lmat{\bs{q}_{\text{b}}} V\hT \bs{p}_{\text{b}}\right),
\end{equation}
where $\bs{p}$ is a vector from the center of mass of a body to the joint. Our algorithm enforces the constraint $\bs{g} = \bs{0}$ at the position level to guarantee constraint satisfaction, whereas other maximal-coordinate methods only enforce $\bs{g}$ at the acceleration level which results in constraint drift \cite{baraff_linear-time_1996}. 

\section{Variational Integrator} \label{sec:variational}
We derive a first-order variational integrator to discretize rigid body dynamics while maintaining realistic energy and momentum conservation behavior. While this derivation is generally not new and has been treated rigorously \cite{marsden_discrete_2001,junge_discrete_2005,manchester_quaternion_2016}, we provide a derivation in maximal coordinates with unified notation. 

Variational integrators are based on the \textit{principle of least action} which states that a mechanical system always takes the path of least action when going from a fixed start point to a fixed end point. Action has the dimensions $[\text{Energy}]\times[\text{Time}]$ and the unforced action integral $S_0$ is defined as 
\begin{equation}\label{eqn:unforceAction}
S_0 = \int_{t_1}^{t_N} \mathcal{L}(\mathbf{x}(t),\dot{\mathbf{x}}(t)) \text{ d}t,
\end{equation}
where $\mathbf{x}$ is a system's trajectory in arbitrary coordinates and $\mathcal{L} = \mathcal{T}-\mathcal{V}$ is the Lagrangian of the system with kinetic energy $\mathcal{T}$ and potential energy $\mathcal{V}$. For the translational case and assuming only gravitational potential, we have
\begin{align}
\mathcal{T}_{\text{T}} &= \ddfrac{1}{2} \bs{v}\hT M \bs{v},\\
\mathcal{V}_{\text{T}} &= g \btx{e}{z}\hT M \bs{x},
\end{align}
with position $\bs{x}$, velocity $\bs{v}$, mass matrix $M = mI_3$, gravitational acceleration $g$, and the z-axis vector $\btx{e}{z}$.
In the rotational setting we assume zero potential, so $\mathcal{V}_{\text{R}} = 0$, and the kinetic energy is
\begin{equation}
\mathcal{T}_{\text{R}} = \ddfrac{1}{2} \bs{\omega}\hT J \bs{\omega}.
\end{equation}

\subsection{Translational Component}
Inserting the Lagrangian for the translational case into \eqref{eqn:unforceAction} and appending integrals for external forces $\bs{F}$ and constraints $\bs{g}$ with virtual constraint forces $\bs{\lambda}$ yields the forced and constrained action integral
\begin{equation}\label{eqn:ST}
S_{\text{T}}(\bs{x}) = \int_{t_1}^{t_N}\left(\ddfrac{1}{2}\bs{v}\hT M\bs{v} - g \btx{e}{z}\hT M \bs{x}\right) \text{ d}t + \int_{t_1}^{t_N}\bs{F}\hT \bs{x} \text{ d}t + \int_{t_1}^{t_N}\bs{\lambda}\hT \bs{g}(\bs{x}) \text{ d}t.
\end{equation}
The virtual constraint forces $\bs{\lambda}$ act on a rigid body to guarantee satisfaction of constraints $\bs{g}(\bs{x})$. They are called virtual as they should be workless and not change the energy of a body.

For numerical integration, we need to discretize \eqref{eqn:ST}. We use a first-order discretization of the integral and a first-order approximation of the velocity, 
\begin{equation}\label{eqn:vk}
\vno{k} = \ddfrac{\xno{k+1}-\xno{k}}{\dt} ,
\end{equation}
to obtain the forced and constrained discrete action sum:
\begin{equation}
S_{\text{d},\text{T}}(\bs{x}) = \sum_{k=1}^{N-1}\left( \ddfrac{1}{2}\vno{k}\hT M\vno{k} - g \btx{e}{z}\hT M \xno{k} + \bno{F}{k}\hT \xno{k} + \lano{k}\hT \bs{g}(\xno{k})\right) \dt
\end{equation}

With our discrete action sum we can now analyze a very short trajectory from, $t_1$ to $t_3$, i.e. $k=1$ to $k=2=N-1$, that satisfies the principle of least action. Since $\xno{1}$ and $\xno{3}$ are fixed start and end points, only $\xno{2}$ can vary. We therefore minimize the discrete action sum with respect to the position $\xno{2}$:
\begin{equation}\label{eqn:derivDiscTrans}
\nabla_{\xno{2}}S_{\text{d},\text{T}}(\xno{2}) =\left(\ddfrac{1}{\dt} M \vno{1} - \ddfrac{1}{\dt} M \vno{2} -g M \btx{e}{z} + \bno{F}{2} + G_{\bs{x}}(\xno{2})\hT \lano{2} \right)\dt = \bs{0},
\end{equation}
where $G_{\bs{x}}$ is the Jacobian of constraints $\bs{g}$ with respect to $\bs{x}$. In words, if equation \eqref{eqn:derivDiscTrans} is fulfilled, we have found the physically correct trajectory $\bs{x}(t)$ with fixed start point $\bs{x}(t_1)$ and fixed end point $\bs{x}(t_3)$.

Rearranging \eqref{eqn:derivDiscTrans} yields the \textit{discretized translational equations of motion} for a single rigid body with forcing and constraints:
\begin{equation}\label{eqn:dT}
\btx{d}{T}(\vno{2}, \lano{2}) = M\left( \ddfrac{\vno{2}-\vno{1}}{\dt} + g\btx{e}{z} \right)-\bno{F}{2} - G_{\bs{x}}(\xno{2})\hT\lano{2} = \bs{0}
\end{equation}

To derive the physically accurate equations of motion we have varied $\xno{2}$ and assumed fixed start and end points $\xno{1}$ and $\xno{3}$. However, when we want to integrate our dynamics forward in time, we are actually trying to find $\xno{3}$ given $\xno{1}$ and $\xno{2}$. The new position $\xno{3}$ can be derived from \eqref{eqn:vk} as
\begin{equation} \label{eqn:updatePos}
\xno{3} = \xno{2} + \vno{2}\dt,
\end{equation}
which means, for simulations, we will have to solve \eqref{eqn:dT} implicitly for $\vno{2}$ while also finding appropriate constraint forces $\lano{2}$.

Higher-order variational integrators can be derived by using higher-order quadrature rules in the discrete action sum to better approximate the true action integral \cite{wenger_construction_2017,ober-blobaum_construction_2015}. While our algorithm is compatible with higher-order integrators, for clarity we are restricting the discussion in this paper to first-order integrators.

\subsection{Rotational Component}\label{sec:varrot}
The rotational component of the integrator can be derived similarly to the translation case. Details on the derivation can be found in the supplementary material and the literature \cite{manchester_quaternion_2016}. We will only state important equations and results here.

The forced and constrained action integral for rotations is
\begin{equation}
S_{\text{R}}(\bs{q}) = \int_{t_1}^{t_N}\ddfrac{1}{2}\bs{\omega}\hT J\bs{\omega} \text{ d}t + \int_{t_1}^{t_N}2\bs{\tau}\hT V \LTmat{\bs{q}} \bs{q} \text{ d}t + \int_{t_1}^{t_N}\bs{\lambda}\hT \bs{g}(\bs{q}) \text{ d}t,
\end{equation}
with torque $\bs{\tau}$, which leads to the discrete forced and constrained action sum
\begin{equation}\label{eqn:discsumrot}
S_{\text{d},\text{R}}(\bs{q}) = \sum_{k=1}^{N-1}\left( \ddfrac{1}{2}\wno{k}\hT J\wno{k} + 2\bs{\tau}_k\hT V \LTmat{\qno{k}} \qno{k} + \lano{k}\hT \bs{g}(\qno{k})\right) \dt.
\end{equation}
In this discrete action sum we have approximated $\dot{\bs{q}}$ as
\begin{equation}\label{eqn:quatapprox}
\dot{\bs{q}} = \ddfrac{\qno{k+1}-\qno{k}}{\dt}.
\end{equation}

To derive the rotational equations of motion, similarly to \eqref{eqn:derivDiscTrans}, we again minimize the discrete action sum from $k=1$ to $k=2$, this time while varying the orientation $\qno{2}$:
\begin{equation}\label{eqn:derivDiscRot}
\nabla^{\text{r}}_{\qno{2}}S_{\text{d},\text{R}}(\bs{q}) = V \LTmat{\qno{2}} \nabla_{\qno{2}}S_{\text{d},\text{R}}(\bs{q}) = \bs{0}
\end{equation}
Note the \textit{rotational gradient} $\nabla^{\text{r}}$ and see the supplementary material for details.

Introducing an implicit unit norm constraint on the quaternions \cite{manchester_quaternion_2016} to avoid explicit constraints yields an update rule for the orientation similar to \eqref{eqn:updatePos}:
\begin{equation} \label{eqn:updateOr}
\qno{3} = \ddfrac{\dt}{2}\Lmat{\qno{2}}\begin{bmatrix}\sqrt{\left(\tfrac{2}{\dt}\right)^2 - \wno{2}\hT\wno{2}}\\ \wno{2}\end{bmatrix}
\end{equation}

With this implicit constraint we can state the \textit{discretized rotational equations of motion}:
\begin{align}\label{eqn:dR}
\begin{split}
\btx{d}{R}(\wno{2}, \lano{2}) &=  J \wno{2}\sqrt{\tfrac{4}{\dt^2}-\wno{2}\hT\wno{2}} + \wno{2}^{\times} J \wno{2}\\ &~~~~ -  J \wno{1}\sqrt{\tfrac{4}{\dt^2}-\wno{1}\hT\wno{1}} + \wno{1}^{\times} J \wno{1} - 2\bno{\tau}{2} -  G_{\bs{q}}(\qno{2})\hT\lano{2}  = \bs{0}.
\end{split}
\end{align}

\section{Linear-Time Sparse Solver}\label{sec:solver}
We can use the discretized equations of motion from Sec.  \ref{sec:variational} to simulate the rigid body dynamics forward in time. For a single rigid body, we stack our equations of motion $\bs{d} = [\bs{d}_{\text{T}}\hT ~~ \bs{d}_{\text{R}}\hT]\hT$ and the constraints $\bs{g}$ into a function
\begin{equation}
\bs{f} = \begin{bmatrix}
\bs{d} \\ \bs{g}
\end{bmatrix}.
\end{equation}
To treat multiple rigid bodies, connected or not, their equations are simply stacked into $\bs{f}$. For example, a pendulum with $n$ links (6 equations of motion each) and $n$ revolute joints (5 constraints each) would result in $\text{dim}(\bs{f})=6n + 5n$. Hence, from now on, $\bs{f}$ contains all equations of motion and constraints for all bodies in a simulation. Our goal is then to find the solution to the non-linear equation
\begin{equation} \label{eq:f}
\bs{f}(\bs{s}) = \bs{0},
\end{equation}
where the solution, $\bs{s}$, consists of $\vno{2}$, $\wno{2}$, and $\lano{2}$. Once we have found these values, $\vno{2}$ and $\wno{2}$ can be used to update the mechanism's position and orientation according to \eqref{eqn:updatePos} and \eqref{eqn:updateOr}.

\subsection{Newton's Method and Dense LDU Decomposition}

Newton's method is used to solve the implicit equation \eqref{eq:f}:\begin{align}
\bs{s}\itind{i+1} &= \bs{s}\itind{i} - \left(F(\bs{s}\itind{i})\right)^{-1} \bs{f}(\bs{s}\itind{i}) \label{eqn:newton}\\
&= \bs{s}\itind{i} - \Delta \bs{s}\itind{i}, 
\end{align}
where $\bs{s}\itind{i}$ is the approximation of the solution at iteration step $i$ and $F$ is the Jacobian of $\bs{f}$ with respect to $\bs{s}\itind{i} = [\vno{2}\hT ~~ \wno{2}\hT ~~ \lano{2}\hT]\hT$:
\begin{equation}\label{eqn:F}
F = \derivb{\bs{f}(\vno{2},\wno{2}, \lano{2})}{(\vno{2},\wno{2}, \lano{2})} = \begin{bmatrix}
\derivb{\bs{d}(\vno{2},\wno{2}, \lano{2})}{(\vno{2},\wno{2})} &~~ \derivb{\bs{d}(\vno{2},\wno{2}, \lano{2})}{\lano{2}}\\
\derivb{\bs{g}(\vno{2},\wno{2},\lano{2})}{(\vno{2},\wno{2})} &~~ \derivb{\bs{g}(\vno{2},\wno{2},\lano{2})}{\lano{2}} 
\end{bmatrix}
=
\begin{bmatrix}
D_{\bs{v},\bs{\omega}} &~~ G_{\bs{x},\bs{q}}\hT\\
G_{\bs{v},\bs{\omega}} &~~ 0
\end{bmatrix}
\end{equation}

For computational efficiency, instead of directly inverting the Jacobian in \eqref{eqn:newton}, one solves the linear system
\begin{equation}\label{eqn:newtonSys}
F(\bs{s}\itind{i}) \Delta \bs{s}\itind{i} = \bs{f}(\bs{s}\itind{i}).
\end{equation}
For a dense matrix $F$ with few non-zeros, a linear system of equations like \eqref{eqn:newtonSys} is typically solved by using a matrix decomposition. Foreshadowing the linear-time algorithm, we review the LDU decomposition \cite{kwak_linear_2004}.

The LDU decomposition, which is essentially a scaled version of Gaussian elimination, provides a numerically robust method to factorize and solve non-symmetric linear systems of equations such as \eqref{eqn:newtonSys}. The factorization part of an LDU decomposition processes a matrix diagonally from top-left to bottom-right. Algorithm \ref{alg:densefactor} describes an in-place factorization (i.e. replacing current entries of the matrix with new factorized values), where $F_{i,j}$ denotes the entry of $F$ in row $i$ and column $j$.

\begin{valgorithm} 
	\begin{algorithmic}[1]
		\caption{Dense In-Place LDU Factorization. Complexity $O(n^3)$.}\label{alg:densefactor}
		\For{$n = 1:\text{length}(\Delta\bs{s})$}
		\For{$i = 1:n-1$}\Comment{L and U factorization}
		\For{$j = 1:i-1$}
		\State $F_{n,i} \leftarrow F_{n,i} - F_{n,j}F_{j,j}F_{j,i}$\Comment{L}
		\State $F_{i,n} \leftarrow F_{i,n} - F_{i,j}F_{j,j}F_{j,n}$\Comment{U}
		\EndFor
		\State $F_{n,i} \leftarrow F_{n,i}F_{i,i}^{-1}$\Comment{L}
		\State $F_{i,n} \leftarrow F_{i,i}^{-1}F_{i,n}$\Comment{U}
		\EndFor
		
		\For{$j = 1:n-1$}\Comment{D factorization}
		\State $F_{n,n} \leftarrow F_{n,n} - F_{n,j}F_{j,j}F_{j,n}$\Comment{D}
		\EndFor
		\EndFor
	\end{algorithmic}
\end{valgorithm}

With this factorization of complexity $O(n^3)$, we have calculated a lower-triangular matrix ``L'', a diagonal matrix ``D'', and an upper-triangular matrix ``U'', with their product being LDU $=F$. The structures of matrices L, D, and U allow for back-substitution $\Delta\bs{s}=\text{U}^{-1}\text{D}^{-1}\text{L}^{-1}\bs{f}$ with complexity $O(n^2)$ as described in Algorithm \ref{alg:densesolve}.

\begin{valgorithm} 
	\begin{algorithmic}[1]
		\caption{Dense LDU Back-Substitution. Complexity $O(n^2)$.}\label{alg:densesolve}
		\State $\Delta\bs{s} \leftarrow \bs{f}$
		\For{$n = 1:\text{length}(\Delta\bs{s})$}\Comment{L back-substitution}
		\For{$j = 1:n-1$}
		\State $\Delta\bs{s}_n \leftarrow \Delta\bs{s}_n - F_{n,j}\Delta\bs{s}_j$\Comment{L}
		\EndFor
		\EndFor
		\For{$n = \text{length}(\Delta\bs{s}):1$}\Comment{D and U back-substitution}
		\State $\Delta\bs{s}_n \leftarrow F_{n,n}^{-1}\Delta\bs{s}_n$\Comment{D}
		\For{$j = n+1:\text{length}(\Delta\bs{s})$}\Comment{U}
		\State $\Delta\bs{s}_n \leftarrow \Delta\bs{s}_n - F_{n,j}\Delta\bs{s}_j$
		\EndFor
		\EndFor
	\end{algorithmic}
\end{valgorithm}

However, the Jacobian $F$ in our problem is sparse and has structural properties that enable factorization and back-substitution of \eqref{eqn:newtonSys} with $O(n)$ complexity.

\subsection{Sparse Factorization and Back-Substitution}
While a linear-time algorithm to solve classical continuous-time dynamics in maximal coordinates has been known for over two decades \cite{baraff_linear-time_1996}, it cannot be applied to variational integrators. The main difference between these two settings lies in the system \eqref{eqn:newtonSys}. While the continuous-time system is symmetric, allowing the use of a modified LDL$\hT$ decomposition, \eqref{eqn:F} is not symmetric (not even block-symmetric) since $D_{\bs{v},\bs{\omega}}$ is not symmetric and generally $G_{\bs{x},\bs{q}} \neq G_{\bs{v},\bs{\omega}}$. These circumstances prevent us from applying the continuous-time method.

For the following analysis, we will treat $F$ from \eqref{eqn:F} as a block matrix. Each block on the diagonal of $F$ represents a body or a constraint of the underlying mechanism and will be called a node. Each off-diagonal block represents a connection between two nodes, i.e. between a body and a constraint. The $i$-th node has its diagonal block denoted by $D_i$, where $D_i=D_{\bs{v},\bs{\omega}}$ for a body and  $D_i=0$ for a constraint. An off-diagonal block connecting constraint $i$ and body $j$ is denoted by $c_{ij}$ for blocks of $G_{\bs{x},\bs{q}}$ or $c'_{ij}$ for blocks of $G_{\bs{v},\bs{\omega}}$.

While the block entries of $F$ are not symmetric, the \textit{sparsity pattern} of zero and non-zero blocks is symmetric due to matching patterns of $G_{\bs{x},\bs{q}}$ and $G_{\bs{v},\bs{\omega}}$. This means that $c_{ij}$ and $c'_{ij}$ are either both zero or both non-zero. We can give a graph representing this pattern that is undirected due to the symmetry and acyclic for loop-free mechanisms. An important insight is that the graph of the pattern of $F$ and the graph of the underlying mechanical structure are identical. An exemplary mechanism and its graph are given in Fig. \ref{fig:mechanism}.

\begin{vfigure}[!htp]
	\centering
	{\scalefont{1.0}
		\begin{tikzpicture}
		\node at (-1.5,0) {(a)};
		
		\draw[fill=black!5,rounded corners=6pt]
		(0,0.0) rectangle ++(2,-0.4);
		\draw[fill=black!5,rounded corners=6pt,rotate=-45]
		(0.2,0.325) rectangle ++(0.4,-1.8);
		\draw[fill=black!5,rounded corners=6pt,rotate=10]
		(-1.015,-0.86) rectangle ++(0.4,-1.4);
		\draw[fill=black!5,rounded corners=6pt,rotate=35]
		(-0.87,-0.32) rectangle ++(0.4,-1.4);
		\draw[fill=black!5,rounded corners=6pt,rotate=15]
		(1.3,-0.4) rectangle ++(0.4,-1.4);
		
		\draw[fill=black] (0.35,-0.2) circle (0.065);
		\draw[fill=black] (-0.62,-1.17) circle (0.065);
		\draw[fill=black] (-0.25,-0.8) circle (0.065);
		\draw[fill=black] (1.6,-0.2) circle (0.065);
		
		\node at (1,-0.2) {1};
		\node at (0.1,-0.5) {2};
		\node at (-0.55,-1.7) {3};
		\node at (0.05,-1.2) {4};
		\node at (1.75,-0.7) {5};
		
		\draw (0.35,-0.2) -- (-0.27,-0.2);
		\draw (-0.62,-1.17) -- (-0.92,-0.85);
		\draw (-0.25,-0.8) -- (-0.55,-0.55);
		\draw (1.6,-0.2) -- (2.17,-0.2);
		
		\node at (-0.4,-0.2) {6};
		\node at (-1.05,-0.75) {7};
		\node at (-0.7,-0.45) {8};
		\node at (2.32,-0.2) {9};

		\node at (4,0) {(b)};
		
		\draw (6.2,-0.2) -- (5.5,-0.5);
		\draw (5.5,-0.45) -- (5.1,-1.2);
		\draw (5.3,-1.1) -- (4.5,-1.4);
		\draw (4.5,-1.4) -- (4.3,-2);
		\draw (5.2,-1.1) -- (6.1,-1.4);
		\draw (6.2,-1.4) -- (6.5,-1.9);
		\draw (6.2,-0.2) -- (7.3,-0.6);
		\draw (7.4,-0.6) -- (7.9,-1.3);
		
%

		\draw[fill=red!30] (6,0.0) rectangle ++(0.5,-0.5);
		\draw[fill=yellow!30] (5,-0.9) rectangle ++(0.5,-0.5);
		\draw[fill=green!30] (4.1,-1.8) rectangle ++(0.5,-0.5);
		\draw[fill=green!30] (6.3,-1.8) rectangle ++(0.5,-0.5);
		\draw[fill=yellow!30] (7.5,-1.1) rectangle ++(0.5,-0.5);
		
		\draw[fill=red!30] (5.5,-0.45) circle (0.275);
		\draw[fill=yellow!30] (4.5,-1.4) circle (0.275);
		\draw[fill=yellow!30] (6.1,-1.4) circle (0.275);
		\draw[fill=red!30] (7.3,-0.6) circle (0.275);
		
		\node at (6.25,-0.25) {1};
		\node at (5.25,-1.15) {2};
		\node at (4.35,-2.05) {3};
		\node at (6.55,-2.05) {4};
		\node at (7.75,-1.35) {5};
		\node at (5.5,-0.45) {6};
		\node at (4.5,-1.4) {7};
		\node at (6.1,-1.4) {8};
		\node at (7.3,-0.6) {9};
		\end{tikzpicture}}
	\caption{(a) A mechanism with five links and four joints. (b) A graph representing the mechanism and its matrix. Squares represent links, circles represent joints. Coloring represents different levels of the tree-shaped graph.}
	\label{fig:mechanism}
\end{vfigure}

We can factorize a matrix associated with the graph in Fig. \ref{fig:mechanism} (b) efficiently by traversing the graph from the leaves to the root. Leaves are characterized by having only a single connection. The root of an acyclic graph can be chosen arbitrarily. Only if we formulate a procedure that follows the structure of the graph in this order during factorization, will we avoid \textit{fill-ins}, i.e. we will not change any zero blocks \cite{duff_direct_2017}. Given the asymmetric entries but symmetric pattern of \eqref{eqn:F}, we can modify the LDU decomposition to develop a linear-time algorithm.

If we reorder the rows and columns of a matrix, we can change the processing order of the LDU factorization. Reordering a matrix from leaves to root according to its graph structure, therefore, creates a factorization without fill-ins. An ordering that creates fill-ins and one that does not are displayed in Fig. \ref{fig:sparsity}.

\begin{vfigure}[!htp]
	\centering
	{\scalefont{1.0}
		\begin{tikzpicture}
		\matrix [matrix of math nodes,
		nodes={rectangle, 
			minimum size=1.2em, text depth=0.25ex,
			inner sep=0pt, outer sep=0pt,
			anchor=center},
		column sep=-0.5\pgflinewidth,
		row sep=-0.5\pgflinewidth,
		inner sep=0pt,
		left delimiter=(, right delimiter=), 
		] (m1)
		{
			D_1\cellcr & 0 & 0 & 0 & 0 & c'_{61} & 0 & 0 & c'_{91}\\
			0 & D_2 & 0 & 0 & 0 & c'_{62} & c'_{72} & c'_{82} & 0\\
			0 & 0 & D_3 & 0 & 0 & 0 & c'_{73} & 0 & 0\\
			0 & 0 & 0 & D_4 & 0 & 0 & 0 & c'_{84} & 0\\
			0 & 0 & 0 & 0 & D_5 & 0 & 0 & 0 & c'_{95}\\
			c_{61} & c_{62} & 0 & 0 & 0 & D_6 & \bullet & \bullet & \bullet\\
			0 & c_{72} & c_{73} & 0 & 0 & \bullet & D_7 & \bullet & \bullet\\
			0 & c_{82} & 0 & c_{84} & 0 & \bullet & \bullet & D_8 & \bullet\\
			c_{91} & 0 & 0 & 0 & c_{95} & \bullet & \bullet & \bullet & D_9\\
		};

		\begin{scope}[on background layer]
		\fill[fill=red!30, rounded corners] 
		(m1-4-9.south west) -- (m1-1-9.south west) -- (m1-1-1.south west) -- (m1-1-1.north west) -- (m1-1-9.north east) -- (m1-4-9.south east);
		
		\draw[draw=black!50, rounded corners] 
		(m1-1-6.south west) -- (m1-1-1.south west) -- (m1-1-1.north west) -- (m1-1-9.north east) -- (m1-5-9.east);
		
		\draw[draw=black!50, rounded corners] 
		(m1-1-6.south east)  -- (m1-1-9.south west) -- (m1-4-9.south west);
		
		\fill[fill=yellow!30, rounded corners] 
		(m1-3-8.south west) -- (m1-2-8.south west) -- (m1-2-2.south west) -- (m1-2-2.north west) -- (m1-2-8.north east) -- (m1-3-8.south east);
		
		\draw[draw=black!50, rounded corners] 
		(m1-2-7.south west) -- (m1-2-2.south west) -- (m1-2-2.north west) -- (m1-2-6.north west);
		
		\draw[draw=black!50, rounded corners] 
		(m1-2-6.north east) -- (m1-2-8.north east) -- (m1-4-8.south east);
		
		\draw[draw=black!50, rounded corners] 
		(m1-2-8.south west)  -- (m1-4-8.south west);

		\filldraw[fill=yellow!30, draw=black!50,rounded corners] 
		(m1-5-9.north west) -- (m1-5-5.north west) -- (m1-5-5.south west) -- (m1-5-9.south east) 
		-- (m1-5-9.north east) ;
		
		\filldraw[fill=green!30, draw=black!50,rounded corners] 
		(m1-4-8.north west) -- (m1-4-4.north west) -- (m1-4-4.south west) -- (m1-4-8.south east) 
		-- (m1-4-8.north east) ;
		\filldraw[fill=green!30, draw=black!50,rounded corners] 
		(m1-3-7.north west) -- (m1-3-3.north west) -- (m1-3-3.south west) -- (m1-3-7.south east) 
		-- (m1-3-7.north east) ;
		\end{scope}
		
		\coordinate (rm1) at (m1-5-9);
		\coordinate (rm1p) at ($(rm1)+(0.75,0)$);
		\coordinate (lm1m) at ($(rm1p)+(0.75,0)$);
		\coordinate (lm1) at ($(lm1m)+(1.4,0)$);

		\matrix [right of=lm1,matrix of math nodes,
		nodes={rectangle, 
			minimum size=1.2em, text depth=0.25ex,
			inner sep=0pt, outer sep=0pt,
			anchor=center},
		column sep=-0.5\pgflinewidth,
		row sep=-0.5\pgflinewidth,
		inner sep=0pt,
		left delimiter=(, right delimiter=),
		] (m2)
		{
			D_5 & c'_{95} & 0 & 0 & 0 & 0 & 0 & 0 & 0 \\
			c_{95} & D_9 & 0 & 0 & 0 & 0 & 0 & 0 & c_{91} \\	
			0 & 0 & D_3 & c'_{73} & 0 & 0 & 0 & 0 & 0 \\
			0 & 0 & c_{73} & D_7 & 0 & 0 &	c_{72} & 0 & 0 \\
			0 & 0 & 0 & 0 & D_4 & c'_{84} & 0 & 0 & 0 \\
			0 & 0 & 0 & 0 & c_{84} & D_8 & c_{82} & 0 & 0 \\
			0 & 0 & 0 & c'_{72} & 0 & c'_{82} & D_2 & c'_{62} & 0 \\
			0 & 0 & 0 & 0 & 0 & 0 & c_{62} & D_6 & c_{61} \\	
			0 & c'_{91} & 0 & 0 & 0 & 0 & 0 & c'_{61} & D_1\\
		};

		\begin{scope}[on background layer]
		\fill[fill=red!30, rounded corners] 
		(m2-2-2.north west) --
		(m2-9-2.south west) -- (m2-9-9.south east) -- (m2-9-9.north east) --
		(m2-9-2.north east) -- (m2-2-2.north east);
		\fill[fill=red!30, rounded corners] 
		(m2-8-8.north west) -- (m2-9-8.north west) -- (m2-9-7.north west) -- 
		(m2-9-7.south west) -- (m2-9-9.south east) -- (m2-9-9.north east) -- 
		(m2-9-8.north east) -- (m2-8-8.north east);
		
		\draw[draw=black!50, rounded corners] 
		(m2-2-2.north west) --
		(m2-9-2.south west) -- (m2-9-9.south east) -- (m2-9-9.north east) -- (m2-9-8.north east)
		-- (m2-8-8.north east);
		
		\draw[draw=black!50, rounded corners] 
		(m2-8-8.north west) --
		(m2-9-8.north west) -- (m2-9-2.north east) -- (m2-2-2.north east);
		
		\filldraw[fill=yellow!30,draw=black!50, rounded corners] 
		(m2-1-2.south east) --  (m2-1-2.north east) --(m2-1-1.north west) -- (m2-1-1.south west)
		-- (m2-1-2.south west);
		
		\fill[fill=yellow!30, rounded corners] 
		(m2-4-4.north west) --
		(m2-7-4.south west) -- (m2-7-8.south east) -- (m2-7-8.north east) --
		(m2-7-4.north east) -- (m2-4-4.north east);
		\fill[fill=yellow!30, rounded corners] 
		(m2-6-6.north west) -- (m2-7-6.north west) -- (m2-7-5.north west) -- 
		(m2-7-5.south west) -- (m2-7-7.south east) -- (m2-7-7.north east) -- 
		(m2-7-6.north east) -- (m2-6-6.north east);
		
		\draw[draw=black!50, rounded corners] 
		(m2-4-4.north west) --
		(m2-7-4.south west) -- (m2-7-8.south west);
		
		\draw[draw=black!50, rounded corners] 
		(m2-7-8.south east) -- (m2-7-8.north east) -- (m2-7-6.north east)
		-- (m2-6-6.north east);
		
		\draw[draw=black!50, rounded corners] 
		(m2-6-6.north west) --
		(m2-7-6.north west) -- (m2-7-4.north east) -- (m2-4-4.north east);
		
		\filldraw[fill=green!30,draw=black!50, rounded corners] 
		(m2-3-4.south east) --  (m2-3-4.north east) --(m2-3-3.north west) -- (m2-3-3.south west)
		-- (m2-3-4.south west);
		
		\filldraw[fill=green!30,draw=black!50, rounded corners] 
		(m2-5-6.south east) --  (m2-5-6.north east) --(m2-5-5.north west) -- (m2-5-5.south west)
		-- (m2-5-6.south west);
		\end{scope} 
		
		\draw [->] (rm1p) -- (lm1m);
		\draw [circle] (rm1p);
		
		\node at ($(m1-1-1)-(0.8,0)$) {(a)};
		\node at ($(m2-1-1)-(0.8,0)$) {(b)};
		\end{tikzpicture}}
	\caption{Matrices for the mechanism in Fig. \ref{fig:mechanism} with matching color scheme. Fill-ins indicated as ``$\bullet$''. (a) Unordered matrix creating fill-ins during LDU factorization. (b) Rearranged matrix without fill-ins during LDU factorization. 
	}
	\label{fig:sparsity}
\end{vfigure}

When looking at the nodes on the diagonal of matrix (b) in Fig. \ref{fig:sparsity} from top-left to bottom-right, one can clearly see how the nodes are ordered from leaves to root $D_1$ according to the graph in Fig. \ref{fig:mechanism} (b). By contrast, in matrix (a) the non-leaf node $D_1$ is processed first. Note that generally there are multiple orderings to achieve zero fill-ins, for example in matrix (b), we could change the order of nodes $D_3$ and $D_7$ with $D_4$ and $D_8$, or chose an entirely different root. 

We know that, for a loop-free structure, each link has exactly one parent. Therefore, for a structure with $n$ links, we also have $n$ constraints (or $n-1$ for floating-base systems). Because we are not creating fill-ins during the factorization, we can achieve $O(n)$ complexity by only evaluating the factorization for the $O(n)$ nodes. To find an appropriate matrix ordering, we have to perform a depth-first-search starting from the (arbitrary) root. We store the found nodes in a list with the root as the last element and the last-found node as the first element. This list then contains a valid ordering of nodes. The modified factorization procedure is formalized in Algorithm \ref{alg:factor}.

\begin{valgorithm} 
	\begin{algorithmic}[1]
		\caption{Sparse In-Place LDU Factorization. Complexity $O(n)$.}\label{alg:factor}
		\For{$i \in \text{list}$}\Comment{list from depth-first-search}
		\For{$c \in \text{children}(i)$}\Comment{Modified L and U factorization of Alg. \ref{alg:densefactor}}
		\State $F_{i,c} \leftarrow F_{i,c}F_{c,c}^{-1}$
		\State $F_{c,i} \leftarrow F_{c,c}^{-1}F_{c,i}$
		\EndFor
		\For{$c \in \text{children}(i)$}\Comment{Modified D factorization of Alg. \ref{alg:densefactor}}
		\State $F_{i,i} \leftarrow F_{i,i} - F_{i,c}F_{c,c}F_{c,i}$
		\EndFor
		\EndFor
	\end{algorithmic}
\end{valgorithm}

We will also exploit the fact that we did not create fill-ins during the factorization and that we have $O(n)$ nodes to develop a back-substitution procedure of $O(n)$ complexity. This modified procedure is described in Algorithm \ref{alg:solve}.

\begin{valgorithm} 
	\begin{algorithmic}[1]
		\caption{Sparse LDU Back-Substitution. Complexity $O(n)$.}\label{alg:solve}
		\For{$i \in \text{list}$}\Comment{list from depth-first-search}
		\State $\Delta\bno{s}{i} \leftarrow \bno{f}{i}$
		\For{$c \in \text{children}(i)$}\Comment{Modified L back-substitution of Alg. \ref{alg:densesolve}}
		\State $\Delta\bno{s}{i} \leftarrow \Delta\bno{s}{i} - F_{i,c}\Delta\bno{s}{c}$
		\EndFor
		\EndFor
		\For{$i \in \text{reverse}(\text{list})$}\Comment{Modified D and U back-substitution of Alg. \ref{alg:densesolve}}
		\State $\Delta\bno{s}{i} \leftarrow F_{i,i}^{-1}\Delta\bno{s}{i}$
		\State $\Delta\bno{s}{i} \leftarrow \Delta\bno{s}{i} - F_{c,\text{par}(i)}\Delta\bno{s}{\text{par}(i)}$ \Comment{par($i$): parent($i$); if $i$ has a parent}
		\EndFor
	\end{algorithmic}
\end{valgorithm}

We perform sparse factorization and back-substitution iteratively until our Newton method converges, which typically takes three to four iterations with a backtracking line search to ensure the residual decreases. Once converged, we extract $\vno{2}$ and $\wno{2}$ from our solution $\bs{s}$ to update $\xno{3}$ and $\qno{3}$ according to \eqref{eqn:updatePos} and \eqref{eqn:updateOr}. Subsequently, we use $\bs{s}$ as the initial guess for the next time step.

\subsection{Extension to Closed-Loop Mechanisms}
The presented algorithms achieve linear-time complexity for loop-free structures only. If we introduce constraints that form loops in the mechanism, we cannot avoid creating fill-ins during the factorization. Deriving an optimal factorization, for example with the minimal possible number of fill-ins, goes beyond the scope of this paper. Nonetheless, we can give a simple strategy to extend our sparse solver to detect and handle loop-closure constraints. 

Mathematically, we treat closed-loop mechanisms just as loop-free ones. This is a considerable advantage for users as no adjustments are required. Algorithmically, loop-closure constraints can be found with the already deployed depth-first-search. We stack all constraints that enforce loop closure into a single node and append it to the end of our depth-first-search list. In this manner, we can process the entire system with the same algorithms as before except for the very last node. For this node, we will perform the standard dense LDU factorization and back-substitution of Algorithms \ref{alg:densefactor} and \ref{alg:densesolve}. If there are only a few of these additional constraints, the majority of our algorithm can still perform in linear-time and only the processing of the last node containing all loop-closure constraints requires additional computational effort.

\section{Experiments and Comparison} \label{sec:results}
We have implemented our algorithm in the programming language Julia \cite{bezanson_julia_2017}. It has been recently shown that Julia can achieve performance comparable to state-of-the-art C++ dynamics implementations \cite{koolen_julia_2019}. The code for our algorithm and all experiments is available at: \href{https://github.com/RoboticExplorationLab}{https://github.com/RoboticExplorationLab}. All experiments were performed on an ASUS ZenBook with an Intel i7 processor.

\subsection{Energy Conservation, No Constraint Drift, and Linear Time}
To validate the main properties of our algorithm---energy conservation, elimination of constraint drift, and $O(n)$ complexity---we ran three representative experiments demonstrating qualitative behavior in Fig. \ref{fig:validation} where we also show typical convergence behavior.

\begin{vfigure}[!htp] 
	\centering
	\begin{subfigure}{0.48\textwidth}
		\resizebox{\linewidth}{!}{\input{plots/energy.tex}}
	\end{subfigure}
	\begin{subfigure}{0.48\textwidth}
		\resizebox{\linewidth}{!}{\input{plots/drift.tex}}
	\end{subfigure}
	\begin{subfigure}{0.48\textwidth}
		\resizebox{\linewidth}{!}{
%
%
\definecolor{mycolor1}{rgb}{0.00000,0.44700,0.74100}%
\definecolor{mycolor2}{rgb}{0.85000,0.32500,0.09800}%
\begin{tikzpicture}

\begin{axis}[%
width=1.91in,
height=1.0in,
at={(0.38in,0.352in)},
scale only axis,
xmin=1,
xmax=10,
xlabel style={font=\color{black}},
xlabel={Number of Links},
ymin=0,
ymax=1.5,
ylabel style={font=\color{black}},
ylabel={Comput. Time (ms)},
xlabel style={yshift = {4},},
ylabel style={yshift = {-12},},
title={(c) \textbf{Timing of Newton Step}},
title style = {yshift = -5,},
axis background/.style={fill=white},
legend style={at={(0.03,0.97)}, anchor=north west, legend cell align=left, align=left, draw=white!15!black}
]
\addplot [color=mycolor1, line width=2.0pt]
  table[row sep=crcr]{%
1	0.005564565\\
2	0.012280965\\
3	0.018174475\\
4	0.023727425\\
5	0.031525455\\
6	0.037623905\\
7	0.043362955\\
8	0.047885985\\
9	0.05273948\\
10	0.05482399\\
};
\addlegendentry{Sparse Solver}

\addplot [color=mycolor2, line width=2.0pt]
  table[row sep=crcr]{%
1	0.0008\\
2	0.058401\\
3	0.141199\\
4	0.2753\\
5	0.441501\\
6	0.624901\\
7	0.8105\\
8	0.958\\
9	1.21\\
10	1.24\\
};
\addlegendentry{Dense Inverse}

\end{axis}

\end{tikzpicture}%
}
	\end{subfigure}
	\begin{subfigure}{0.48\textwidth}
		\resizebox{\linewidth}{!}{
%
%
\definecolor{mycolor1}{rgb}{0.75098,0.22549,0.59804}%
\definecolor{mycolor2}{rgb}{0.92941,0.69412,0.12549}%
\definecolor{mycolor3}{rgb}{0.00000,0.49804,0.00000}%
\begin{tikzpicture}

\begin{axis}[%
width=1.775in,
height=1.094in,
at={(0.516in,0.352in)},
scale only axis,
unbounded coords=jump,
xmin=0,
xmax=4,
xlabel style={font=\color{black}},
xlabel={Iterations},
ymode=log,
ymin=1e-17,
ymax=100000,
ytick={1e-15,1e-10,1e-05,1,100000},
yticklabels={{$10^{-15}$},{$10^{-10}$},{$10^{-5}$},{$10^{0}$},{$10^{5}$}},
yminorticks=true,
ylabel style={font=\color{black}},
ylabel={Solution Accuracy},
xlabel style={yshift = {4},},
title={(d) \textbf{Convergence Behavior}},
title style = {yshift = -5,},
axis background/.style={fill=white},
legend style={legend cell align=left, align=left, draw=white!15!black}
]
\addplot [color=mycolor1, line width=2.0pt, mark=x, mark options={solid, mycolor1}]
  table[row sep=crcr]{%
0	9.81\\
1	3.57e-07\\
2	1.21e-15\\
3	5e-16\\
4	nan\\
};
\addlegendentry{1 Link}

\addplot [color=mycolor2, line width=2.0pt, mark=x, mark options={solid, mycolor2}]
  table[row sep=crcr]{%
0	31.02194385\\
1	3.16e-07\\
2	1.78e-14\\
3	1.63e-14\\
4	nan\\
};
\addlegendentry{10 Links}

\addplot [color=mycolor3, line width=2.0pt, mark=x, mark options={solid, fill=mycolor3, mycolor3}]
  table[row sep=crcr]{%
0	98.1\\
1	3.16e-07\\
2	4.29e-13\\
3	4.56e-13\\
4	nan\\
};
\addlegendentry{100 Links}

\addplot [color=black, dashed, line width=1.0pt, forget plot]
  table[row sep=crcr]{%
0	1e-10\\
1	1e-10\\
2	1e-10\\
3	1e-10\\
4	1e-10\\
};
\end{axis}
\end{tikzpicture}%
}
	\end{subfigure}
	\caption{Main algorithm properties. Our algorithm in blue, comparison in red. (a) Energy conservation with variational integrator compared to explicit integrator. (b) Drift with position constraints compared to drift with acceleration constraints. (c) Sparse $O(n)$ behavior compared to $O(n^3)$ dense matrix inversion. (d) Convergence behavior of Newton's method in our algorithm.}\label{fig:validation}
\end{vfigure} 

For plot (a) we measured the energy of a double pendulum over 60 minutes with a time step $\dt = 10\text{ms}$. We compare our first-order variational integrator to the explicit second-order Heun's method (a Runge-Kutta method), using the \textit{DifferentialEquations.jl} package \cite{rackauckas_differentialequations.jl_2017}. The error increases substantially over time with the explicit method, a behavior typical for (lower-order) explicit integrators, whereas the energy stays constant over time with our variational approach.

To determine drift behavior, for plot (b) we simulated a four-link closed-loop mechanism with our maximal-coordinate approach and compared it to a state-of-the-art minimal-coordinate dynamics package, \textit{RigidBodyDynamics.jl} \cite{koolen_julia_2019}. The loop closure introduces constraints to the minimal-coordinate approach. As the minimal-coordinate approach only enforces acceleration constraints, without constraint stabilization, an increasing drift becomes visible. In contrast, our algorithm directly enforces position constraints, which guarantees constraint satisfaction without requiring any additional constraint stabilization. 

Plot (c) compares the computation time of sparse and dense single Newton steps for mechanisms with varying numbers of links. The rapid increase in computation time for a larger number of links when using dense inversion compared to our sparse approach highlights the necessity to exploit sparsity to achieve good performance. 

We demonstrate quadratic convergence behavior, i.e. doubling accuracy at each iteration step, for simulating a single time step of pendulums with 1, 10, and 100 links in plot (d). The termination condition is a converged solution $\lVert \Delta \bs{s}\itind{i+1}-\Delta \bs{s}\itind{i}\rVert < \epsilon$ with a tolerance of $\epsilon = 10^{-10}$.

\subsection{Performance Comparison}
We compare the performance of our algorithm to three minimal-coordinate implementations: the \textit{RigidBodyDynamics.jl} package with a state-of-the-art explicit Runge-Kutta-Munthe-Kaas integrator \cite{hairer_geometric_2006}, and two C++ implementations of variational integrators in minimal coordinates using quasi-Newton methods \cite{lee_linear-time_2016,fan_efficient_2018}. We ran all algorithms with the same initial conditions over 1000 time steps to cover a reasonable area of the state space of each mechanism. The best timing result of 100 runs was taken to eliminate right-skewing computer noise.

\subsubsection{Comparison with Explicit Integrator}
For the comparison with the explicit integrator, we simulated swinging pendulums---having snake robots \cite{tesch_parameterized_2009} or finite-element-method soft robotics \cite{duriez_control_2013} in mind---with varying numbers of links for 10
\begin{vfigure}[!htp] 
	\centering
	\begin{subfigure}{0.48\textwidth}
		\resizebox{\linewidth}{!}{
%
%
\definecolor{mycolor1}{rgb}{0.00000,0.44700,0.74100}%
\definecolor{mycolor2}{rgb}{0.00000,0.44706,0.74118}%
\definecolor{mycolor3}{rgb}{0.85098,0.32549,0.09804}%
\begin{tikzpicture}

\begin{axis}[%
width=1.962in,
height=1.33in,
at={(0.329in,0.385in)},
scale only axis,
xmin=1,
xmax=100,
xlabel style={font=\color{black}},
xlabel={Number of Links},
ymin=0,
ymax=5,
ylabel style={font=\color{black}},
ylabel={Comput. Time (s)},
xlabel style={yshift = {4},},
ylabel style={yshift = {-16},},
title={(a) \textbf{Pendulum}},
title style = {yshift = -5,},
axis background/.style={fill=white},
legend style={at={(0.03,0.97)}, anchor=north west, legend cell align=left, align=left, draw=white!15!black, fill opacity=0.8}
]
\addplot [color=mycolor1, line width=2.0pt]
  table[row sep=crcr]{%
1	0.00686\\
2	0.0174\\
3	0.0305\\
4	0.0422\\
5	0.0536\\
6	0.0648\\
7	0.0773\\
8	0.0925\\
9	0.108\\
10	0.125\\
20	0.338\\
30	0.545\\
40	0.707\\
50	0.883\\
60	1.07\\
70	1.26\\
80	1.44\\
90	1.69\\
100	2\\
};
\addlegendentry{Max. Co. Rev.}

\addplot [color=mycolor2, dashed, line width=2.0pt]
  table[row sep=crcr]{%
1	0.00603\\
2	0.0143\\
3	0.0243\\
4	0.0336\\
5	0.044\\
6	0.0523\\
7	0.0623\\
8	0.0752\\
9	0.0878\\
10	0.103\\
20	0.263\\
30	0.438\\
40	0.581\\
50	0.748\\
60	0.846\\
70	0.991\\
80	1.17\\
90	1.39\\
100	1.66\\
};
\addlegendentry{Max. Co. Ball.}

\addplot [color=mycolor3, line width=2.0pt]
  table[row sep=crcr]{%
1	0.00384\\
2	0.00534\\
3	0.00711\\
4	0.00859\\
5	0.0103\\
6	0.0118\\
7	0.0139\\
8	0.0151\\
9	0.0174\\
10	0.0189\\
20	0.0377\\
30	0.0591\\
40	0.308\\
50	0.359\\
60	0.41\\
70	0.662\\
80	0.921\\
90	0.972\\
100	1.05\\
};
\addlegendentry{Min. Co. Rev.}

\addplot [color=mycolor3, dashed, line width=2.0pt]
  table[row sep=crcr]{%
1	0.00535\\
2	0.00837\\
3	0.00986\\
4	0.0152\\
5	0.0162\\
6	0.0199\\
7	0.0279\\
8	0.0317\\
9	0.0322\\
10	0.0362\\
20	0.345\\
30	0.859\\
40	1.04\\
50	1.85\\
60	2.28\\
70	2.64\\
80	2.99\\
90	3.62\\
100	4.26\\
};
\addlegendentry{Min. Co. Ball.}

\end{axis}
\end{tikzpicture}%
}
	\end{subfigure}
	\begin{subfigure}{0.48\textwidth}
		\resizebox{\linewidth}{!}{
%
%
\definecolor{mycolor1}{rgb}{0.00000,0.44700,0.74100}%
\definecolor{mycolor2}{rgb}{0.00000,0.44706,0.74118}%
\definecolor{mycolor3}{rgb}{0.85098,0.32549,0.09804}%
\begin{tikzpicture}

\begin{axis}[%
width=1.962in,
height=1.33in,
at={(0.329in,0.385in)},
scale only axis,
xmin=3,
xmax=101,
xlabel style={font=\color{black}},
xlabel={Number of Links},
ymin=0,
ymax=5,
ylabel style={font=\color{black}},
ylabel={Comput. Time (s)},
xlabel style={yshift = {4},},
ylabel style={yshift = {-16},},
title={(b) \textbf{Closed Chain}},
title style = {yshift = -5,},
axis background/.style={fill=white},
legend style={at={(0.03,0.97)}, anchor=north west, legend cell align=left, align=left, draw=white!15!black, fill opacity=0.8}
]
\addplot [color=mycolor1, line width=2.0pt]
  table[row sep=crcr]{%
3	0.0359\\
5	0.0769\\
7	0.109\\
9	0.14\\
11	0.18\\
21	0.422\\
31	0.715\\
41	0.964\\
51	1.25\\
61	1.44\\
71	1.72\\
81	1.98\\
91	2.22\\
101	2.54\\
};
\addlegendentry{Max. Co. Rev.}

\addplot [color=mycolor2, dashed, line width=2.0pt]
  table[row sep=crcr]{%
3	0.0279\\
5	0.0552\\
7	0.0855\\
9	0.11\\
11	0.131\\
21	0.33\\
31	0.572\\
41	0.771\\
51	0.971\\
61	1.16\\
71	1.36\\
81	1.59\\
91	1.79\\
101	2.04\\
};
\addlegendentry{Max. Co. Ball.}

\addplot [color=mycolor3, line width=2.0pt]
  table[row sep=crcr]{%
3	0.0221\\
5	0.0264\\
7	0.0316\\
9	0.0365\\
11	0.0406\\
21	0.0663\\
31	0.0944\\
41	0.385\\
51	0.458\\
61	0.533\\
71	0.799\\
81	1.06\\
91	1.14\\
101	1.19\\
};
\addlegendentry{Min. Co. Rev.}

\addplot [color=mycolor3, dashed, line width=2.0pt]
  table[row sep=crcr]{%
3	0.026\\
5	0.0358\\
7	0.0462\\
9	0.0581\\
11	0.314\\
21	0.445\\
31	1.01\\
41	1.2\\
51	2.26\\
61	2.62\\
71	2.97\\
81	3.39\\
91	4.11\\
101	4.85\\
};
\addlegendentry{Min. Co. Ball.}

\end{axis}

\end{tikzpicture}%
}
	\end{subfigure}
	\caption{Performance comparison simulating 1000 time steps of structures with different numbers of links. Our algorithm (maximal coordinates) in blue, explicit integrator (minimal coordinates) in red. Revolute joints solid line, ball-and-socket joints dashed line. (a) Pendulums (b) Closed-loop mechanisms.}\label{fig:rbdcomp}
\end{vfigure}
seconds and a time step $\dt=10$ms. We show the computation time for revolute joints (1 degree of freedom) and ball-and-socket joints (3 degrees of freedom) in plot (a) of Fig. \ref{fig:rbdcomp}. We also introduced a single loop closure and repeated the experiment for these closed-loop structures with the results shown in plot (b). 

In both the open and closed-loop experiments, the linear-time behavior is clearly visible. While the explicit integrator performs better for revolute joints, especially with few degrees of freedom, our algorithm achieves competitive results and outperforms the explicit integrator for ball-and-socket joints. Note that our method is faster at simulating ball-and-socket joints than revolute joints because a ball-and-socket joint constrains fewer degrees of freedom. 

To highlight the robustness of our algorithm, we also simulated a chain of four-link closed-loop segments with revolute joints shown in Fig. \ref{fig:rbdcomp2} on the right. 

\begin{vfigure}[!htp] 
	\centering
	\begin{subfigure}{0.48\textwidth}
		\resizebox{\linewidth}{!}{
%
%
\definecolor{mycolor1}{rgb}{0.00000,0.44700,0.74100}%
\begin{tikzpicture}

\begin{axis}[%
width=1.874in,
height=1.094in,
at={(0.417in,0.352in)},
scale only axis,
xmin=4,
xmax=40,
xlabel style={font=\color{black}},
xlabel={Number of Links},
ymin=0,
ymax=1.5,
ylabel style={font=\color{black}},
ylabel={Comput. Time (s)},
xlabel style={yshift = {4},},
ylabel style={yshift = {-12},},
title={\textbf{Segmented Chain}},
title style = {yshift = -5,},
axis background/.style={fill=white},
legend style={at={(0.03,0.97)}, anchor=north west, legend cell align=left, align=left, draw=white!15!black}
]
\addplot [color=mycolor1, line width=2.0pt]
  table[row sep=crcr]{%
4	0.0599\\
8	0.137\\
12	0.222\\
16	0.319\\
20	0.49\\
24	0.589\\
28	0.726\\
32	0.953\\
36	1.12\\
40	1.34\\
};
\addlegendentry{Max. Co. Rev.}

\end{axis}
\end{tikzpicture}%
}
	\end{subfigure}
	\begin{subfigure}{0.3\textwidth}
		\resizebox{\linewidth}{!}{\includegraphics{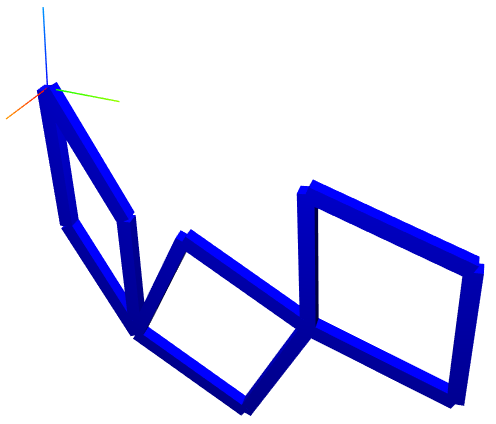}}
	\end{subfigure}
	\caption{Performance simulating 1000 time steps of different chains consisting of four-link segments (example of a chain consisting of three 4-link segments on the right). Timing results for our algorithm.}\label{fig:rbdcomp2}
\end{vfigure} 

\begin{wrapfigure}{l}{0.4\textwidth}
	\vspace*{-8mm}
	\centering
	\resizebox{\linewidth}{!}{\includegraphics{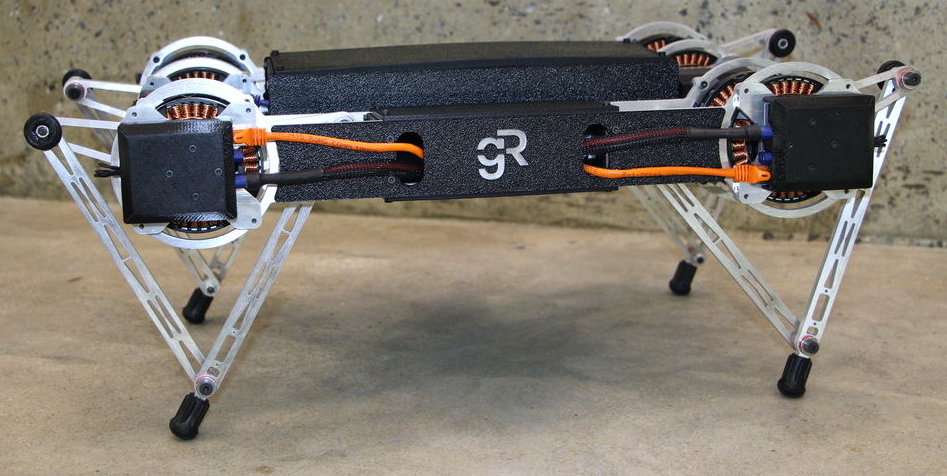}}
	\caption{Quadrupedal robot \textit{Minitaur} \cite{blackman_gait_2016} with closed-loop legs.}
	\vspace*{-4mm}
	\label{fig:walker}
\end{wrapfigure}
Such closed-loop segments can be used as legs for quadrupedal robots, for example \textit{Minitaur} \cite{blackman_gait_2016}, shown in Fig. \ref{fig:walker}. The segments have singular configurations when links overlap. Such kinematic singularities are generally difficult to handle with minimal-coordinate approaches. The results for our algorithm simulating segmented chains of different lengths for 10 seconds with $\dt=10$ms are displayed in Fig. \ref{fig:rbdcomp2} on the left. The minimal-coordinate integrator was unable to reliably simulate chains of two or more four-link segments, and using smaller time steps did not lead to a significant improvement.

\subsubsection{Comparison with Variational Integrators}
As a final experiment, we compared our algorithm to state-of-the-art second- and third-order variational integrators with the same n-link pendulum we used for the comparison to the explicit integrator. Only revolute joints and loop-free structures were tested as ball-and-socket joints and loop-closure were, to the best of our knowledge, not part of the implementations of these integrators. To demonstrate the robustness of our algorithm, we ran simulations with solution tolerances of $10^{-6}$, $10^{-8}$, and $10^{-10}$ for the Newton methods in all algorithms. 
The results are given in Fig. \ref{fig:varcomp}.

\begin{vfigure}[!htp] 
	\centering
	\begin{subfigure}{0.32\textwidth}
		\resizebox{\linewidth}{!}{
%
%
\definecolor{mycolor1}{rgb}{0.00000,0.44700,0.74100}%
\definecolor{mycolor2}{rgb}{0.85000,0.32500,0.09800}%
\definecolor{mycolor3}{rgb}{0.92900,0.69400,0.12500}%
\begin{tikzpicture}

\begin{axis}[%
width=1.359in,
height=1.363in,
at={(0.387in,0.353in)},
scale only axis,
xmin=1,
xmax=100,
xlabel style={font=\color{black}},
xlabel={Number of Links},
ymin=0,
ymax=2.5,
ylabel style={font=\color{black}},
ylabel={Comput. Time (s)},
xlabel style={yshift = {4},},
ylabel style={yshift = {-18},},
title={(a) \textbf{Tolerance of $10^{-6}$}},
title style = {yshift = -5,},
axis background/.style={fill=white},
legend style={at={(0.03,0.97)}, anchor=north west, legend cell align=left, align=left, draw=white!15!black, fill opacity=0.8}
]
\addplot [color=mycolor1, line width=2.0pt]
  table[row sep=crcr]{%
1	0.0075664\\
2	0.01683895\\
3	0.0253815\\
4	0.03037195\\
5	0.04017\\
6	0.045604649\\
7	0.05642305\\
8	0.0673038\\
9	0.072613651\\
10	0.0793415\\
20	0.1869494\\
30	0.328837801\\
40	0.4311182\\
50	0.551548\\
60	0.7164087\\
70	0.8153477\\
80	0.9512933\\
90	1.04255385\\
100	1.198255301\\
};
\addlegendentry{Max. Co. 1$^{\text{st}}$ o.}

\addplot [color=mycolor2, line width=2.0pt]
  table[row sep=crcr]{%
1	0.007249\\
2	0.013966\\
3	0.02021\\
4	0.028086\\
5	0.034534\\
6	0.040594\\
7	0.048243\\
8	0.05384\\
9	0.060522\\
10	0.066505\\
20	0.129724\\
30	0.194994\\
40	0.261957\\
50	0.336895\\
60	0.413523\\
70	0.489115\\
80	0.563091\\
90	0.642634\\
100	0.723301\\
};
\addlegendentry{Min. Co. 2$^{\text{nd}}$ o.}

\addplot [color=mycolor3, line width=2.0pt]
  table[row sep=crcr]{%
1	0.0311637\\
2	0.0418929\\
3	0.0551968\\
4	0.0749877\\
5	0.0836943\\
6	0.0934797\\
7	0.106626\\
8	0.123638\\
9	0.143544\\
10	0.171381\\
20	0.311008\\
30	0.517138\\
40	0.675311\\
50	0.819718\\
60	0.978916\\
70	1.12722\\
80	1.28035\\
90	1.40865\\
100	1.58877\\
};
\addlegendentry{Min. Co. 3$^{\text{rd}}$ o.}

\end{axis}
\end{tikzpicture}%
}
	\end{subfigure}
	\begin{subfigure}{0.32\textwidth}
		\resizebox{\linewidth}{!}{
%
%
\definecolor{mycolor1}{rgb}{0.00000,0.44700,0.74100}%
\definecolor{mycolor2}{rgb}{0.85000,0.32500,0.09800}%
\definecolor{mycolor3}{rgb}{0.92900,0.69400,0.12500}%
\begin{tikzpicture}

\begin{axis}[%
width=1.359in,
height=1.363in,
at={(0.387in,0.353in)},
scale only axis,
unbounded coords=jump,
xmin=1,
xmax=100,
xlabel style={font=\color{white!15!black}},
xlabel={Number of Links},
ymin=0,
ymax=2.5,
ylabel style={font=\color{white!15!black}},
ylabel={Comput. Time (s)},
xlabel style={yshift = {4},},
ylabel style={yshift = {-18},},
title={(b) \textbf{Tolerance of $10^{-8}$}},
title style = {yshift = -5,},
axis background/.style={fill=white},
legend style={at={(0.03,0.97)}, anchor=north west, legend cell align=left, align=left, draw=white!15!black,fill opacity=0.8}
]
\addplot [color=mycolor1, line width=2.0pt]
  table[row sep=crcr]{%
1	0.00712065\\
2	0.01641535\\
3	0.0254385\\
4	0.039339601\\
5	0.053709\\
6	0.061170351\\
7	0.0712449\\
8	0.085226\\
9	0.0960977\\
10	0.10871385\\
20	0.23544205\\
30	0.340059401\\
40	0.4866466\\
50	0.657334201\\
60	0.84990935\\
70	1.161463601\\
80	1.3970568\\
90	1.51375145\\
100	1.7253295\\
};

\addplot [color=mycolor2, line width=2.0pt]
  table[row sep=crcr]{%
1	0.007038\\
2	0.013575\\
3	0.020043\\
4	0.026976\\
5	0.034347\\
6	0.042497\\
7	0.048327\\
8	0.054646\\
9	0.061681\\
10	0.068408\\
20	0.134118\\
30	0.202212\\
40	0.273835\\
50	0.357547\\
60	0.434355\\
70	0.507673\\
80	0.590887\\
86  0.647635\\
90	nan\\
100	nan\\
};

\addplot [color=mycolor3, line width=2.0pt]
  table[row sep=crcr]{%
1	0.0304928\\
2	0.0426332\\
3	0.066099\\
4	0.0875555\\
5	0.106719\\
6	0.141481\\
7	0.158775\\
8	0.173936\\
9	0.19138\\
10	0.207368\\
20	0.316058\\
30	0.520743\\
40	0.67648\\
50	0.757208\\
60	1.20517\\
70	1.94501\\
71  2.02998\\
80	nan\\
90	nan\\
100	nan\\
};

\end{axis}
\end{tikzpicture}%
}
	\end{subfigure}
	\begin{subfigure}{0.32\textwidth}
		\resizebox{\linewidth}{!}{
%
%
\definecolor{mycolor1}{rgb}{0.00000,0.44700,0.74100}%
\definecolor{mycolor2}{rgb}{0.85000,0.32500,0.09800}%
\definecolor{mycolor3}{rgb}{0.92900,0.69400,0.12500}%
\begin{tikzpicture}

\begin{axis}[%
width=1.359in,
height=1.363in,
at={(0.387in,0.353in)},
scale only axis,
unbounded coords=jump,
xmin=1,
xmax=100,
xlabel style={font=\color{black}},
xlabel={Number of Links},
ymin=0,
ymax=2.5,
ylabel style={font=\color{black}},
ylabel={Comput. Time (s)},
xlabel style={yshift = {4},},
ylabel style={yshift = {-18},},
title={(c) \textbf{Tolerance of $10^{-10}$}},
title style = {yshift = -5,},
axis background/.style={fill=white},
legend style={at={(0.03,0.97)}, anchor=north west, legend cell align=left, align=left, draw=white!15!black}
]
\addplot [color=mycolor1, line width=2.0pt]
  table[row sep=crcr]{%
1	0.0068599\\
2	0.017363399\\
3	0.03048355\\
4	0.04224055\\
5	0.0536419\\
6	0.064802901\\
7	0.077306951\\
8	0.092493151\\
9	0.108261151\\
10	0.12507895\\
20	0.33793475\\
30	0.5452531\\
40	0.706500451\\
50	0.882511\\
60	1.0741552\\
70	1.2598156\\
80	1.443597351\\
90	1.6916424\\
100	2.002131901\\
};

\addplot [color=mycolor2, line width=2.0pt]
  table[row sep=crcr]{%
1	0.007269\\
2	0.014203\\
3	0.021192\\
4	0.029606\\
5	0.036335\\
6	0.043272\\
7	0.050918\\
8	0.057401\\
9	0.064768\\
10	0.071678\\
20	0.140861\\
23  0.163760\\
30	nan\\
40	nan\\
50	nan\\
60	nan\\
70	nan\\
80	nan\\
90	nan\\
100	nan\\
};

\addplot [color=mycolor3, line width=2.0pt]
  table[row sep=crcr]{%
1	0.0307483\\
2	0.0620876\\
3	0.0820016\\
4	0.0956382\\
5	0.127227\\
6	0.142604\\
7	0.158592\\
8	0.173507\\
9	0.191319\\
10	0.207469\\
12  0.211639\\
20	nan\\
30	nan\\
40	nan\\
50	nan\\
60	nan\\
70	nan\\
80	nan\\
90	nan\\
100	nan\\
};

\end{axis}
\end{tikzpicture}%
}
	\end{subfigure}
	\caption{Performance comparison simulating 1000 time steps of n-link pendulums with varying solution tolerances. Our ($1^{\text{st}}$ order) algorithm in blue, $2^{\text{nd}}$ order variational integrator in red, $3^{\text{rd}}$ order variational integrator in yellow. (a) Tolerance of $10^{-6}$. (b) Tolerance of $10^{-8}$. (c) Tolerance of $10^{-10}$.}
	\label{fig:varcomp}
\end{vfigure} 

An increasing number of links and lower tolerances lead to failure to converge to a solution for the minimal-coordinate algorithms (stopped after 100 Newton steps). This result demonstrates the numerical robustness and scalability of the direct LDU factorization compared to the recursive calculation in minimal coordinates and is in line with earlier findings \cite{baraff_linear-time_1996}.

\section{Conclusion}
We have presented a novel linear-time variational integrator in maximal coordinates to simulate open and closed-loop mechanical structures without constraint drift. To the best of our knowledge, no previous attempt has been made to develop a linear-time variational integrator in maximal coordinates.

Overall, the results of our experiments show that our algorithm is competitive in terms of computational performance to minimal-coordinate integrators, both explicit and variational. Especially for mechanical systems with a larger number of links and higher-degree-of-freedom joints our method shows better performance for both open and closed-loop structures. Thanks to the numerical stability and robustness of our specialized sparse LDU factorization, our method also achieves tighter numerical tolerances than comparable minimal-coordinate variational integrators. For structures with multiple closed kinematic loops and kinematic singularities, we perform robustly while a minimal-coordinate integrator frequently fails to find solutions.

In summary, the results make a strong case for the use of maximal-coordinate integrators, especially when handling complex, high-degree-of-freedom, closed-loop structures. Additionally, the modular structure of maximal-coordinate approaches opens up the possibility to parallelize computations in the future.

\bibliography{bib}

\begin{thebibliography}{10}

\bibitem{baraff_linear-time_1996}
D.~Baraff.
\newblock Linear-{Time} {Dynamics} {Using} {Lagrange} {Multipliers}.
\newblock In {\em {ACM} {SIGGRAPH} 96}, pages 137--146, 1996.

\bibitem{baumgarte_stabilization_1972}
J.~Baumgarte.
\newblock Stabilization of {Constraints} and {Integrals} of {Motion} in
  {Dynamical} {Systems}.
\newblock {\em Computer Methods in Appl. Mechanics and Engineering},
  1(1):1--16, 1972.

\bibitem{bezanson_julia_2017}
J.~Bezanson, A.~Edelman, S.~Karpinski, and V.~Shah.
\newblock Julia: {A} {Fresh} {Approach} to {Numerical} {Computing}.
\newblock {\em SIAM Review}, 59(1):65--98, 2017.

\bibitem{blackman_gait_2016}
D.~Blackman, J~Nicholson, C~Ordonez, B.~Miller, and J.~Clark.
\newblock Gait {Development} on {Minitaur}, a {Direct} {Drive} {Quadrupedal}
  {Robot}.
\newblock In {\em {SPIE} {Defense}+{Security}}, 2016.

\bibitem{duff_direct_2017}
I.~Duff, A.~Erisman, and J.~Reid.
\newblock {\em Direct {Methods} for {Sparse} {Matrices}}.
\newblock Oxford University Press, 2017.

\bibitem{duriez_control_2013}
C.~Duriez.
\newblock Control of {Elastic} {Soft} {Robots} based on {Real}-{Time} {Finite}
  {Element} {Method}.
\newblock In {\em 2013 {International} {Conference} on {Robotics} and
  {Automation} ({ICRA})}, 2016.

\bibitem{fan_efficient_2018}
T.~Fan, J.~Schultz, and T.~Murphey.
\newblock Efficient {Computation} of {Higher}-{Order} {Variational}
  {Integrators} in {Robotic} {Simulation} and {Trajectory} {Optimization}.
\newblock In {\em Workshop on the {Algorithmic} {Foundations} of {Robotics}
  ({WAFR})}, 2018.

\bibitem{Featherstone08}
R.~Featherstone.
\newblock {\em Rigid {Body} {Dynamics} {Algorithms}}.
\newblock Springer, 2008.

\bibitem{hairer_geometric_2006}
E.~Hairer, C.~Lubich, and G.~Wanner.
\newblock {\em Geometric Numerical Integration}.
\newblock Springer, 2006.

\bibitem{junge_discrete_2005}
O.~Junge, J.~Marsden, and S.~Ober-Bl\"obaum.
\newblock Discrete {Mechanics} and {Optimal} {Control}.
\newblock {\em IFAC Proceedings Volumes}, 38(1):538--543, 2005.

\bibitem{koolen_julia_2019}
T.~Koolen and R.~Deits.
\newblock Julia for {Robotics}: {Simulation} and {Real}-{Time} {Control} in a
  {High}-{Level} {Programming} {Language}.
\newblock In {\em 2019 {International} {Conference} on {Robotics} and
  {Automation} ({ICRA})}, pages 604--611, 2019.

\bibitem{kwak_linear_2004}
J.~Kwak and H.~Sungpyo.
\newblock {\em Linear {Algebra}}.
\newblock Birkh\"auser, 2004.

\bibitem{lee_linear-time_2016}
J.~Lee, C.~Liu, F.~Park, and S.~Srinivasa.
\newblock A {Linear}-{Time} {Variational} {Integrator} for {Multibody}
  {Systems}.
\newblock In {\em Workshop on the {Algorithmic} {Foundations} of {Robotics}
  ({WAFR})}, 2016.

\bibitem{macklin_unified_2014}
M.~Macklin, M.~M\"uller, N.~Chentanez, and T.~Kim.
\newblock Unified {Particle} {Physics} for {Real}-{Time} {Applications}.
\newblock {\em ACM Transactions on Graphics}, 33(4):1--12, 2014.

\bibitem{manchester_quaternion_2016}
Z.~Manchester and M.~Peck.
\newblock Quaternion {Variational} {Integrators} for {Spacecraft} {Dynamics}.
\newblock {\em Journal of Guidance, Control, and Dynamics}, 39(1):69--76, 2016.

\bibitem{marsden_discrete_2001}
J.~Marsden and M.~West.
\newblock Discrete {Mechanics} and {Variational} {Integrators}.
\newblock {\em Acta Numerica}, 10:357--514, 2001.

\bibitem{ober-blobaum_construction_2015}
S.~Ober-Bl\"obaum and N.~Saake.
\newblock Construction and {Analysis} of {Higher} {Order} {Galerkin}
  {Variational} {Integrators}.
\newblock {\em Advances in Computational Mathematics}, 41:955--986, 2015.

\bibitem{rackauckas_differentialequations.jl_2017}
C.~Rackauckas and Q.~Nie.
\newblock {DifferentialEquations}.jl – {A} {Performant} and {Feature}-{Rich}
  {Ecosystem} for {Solving} {Differential} {Equations} in {Julia}.
\newblock {\em Journal of Open Research Software}, 5(1):15, 2017.

\bibitem{siciliano_springer_2016}
B.~Siciliano and O.~Khatib.
\newblock {\em Springer {Handbook} of {Robotics}}.
\newblock Springer, 2016.

\bibitem{sola_quaternion_2017}
J.~Sol\`a.
\newblock Quaternion {Kinematics} for the {Error}-{State} {Kalman} {Filter}.
\newblock {\em arXiv e-prints}, arXiv:1711.02508 [cs.RO], 2017.

\bibitem{tesch_parameterized_2009}
M.~Tesch, K.~Lipkin, I.~Brown, R.~Hatton, A.~Peck, J.~Rembisz, and H.~Choset.
\newblock Parameterized and {Scripted} {Gaits} for {Modular} {Snake} {Robots}.
\newblock {\em Advanced Robotics}, 23(9):1131--1158, 2009.

\bibitem{wenger_construction_2017}
T.~Wenger, S.~Ober-Bl\"obaum, and S.~Leyendecker.
\newblock Construction and {Analysis} of {Higher} {Order} {Variational}
  {Integrators} for {Dynamical} {Systems} with {Holonomic} {Constraints}.
\newblock {\em Advances in Computational Mathematics}, 43(5):1163--1195, 2017.

\end{thebibliography}
\bibliographystyle{plain}

\mainmatter              
\title{Supplementary Material for: Linear-Time Variational Integrators in Maximal Coordinates}
\titlerunning{Linear-Time Variational Integrators in Maximal Coordinates}  
%
\author{Jan Br\"udigam \and Zachary Manchester}
\authorrunning{Jan Br\"udigam and Zachary Manchester} 
%
\tocauthor{Jan Br\"udigam and Zachary Manchester}
\institute{Stanford University, Stanford CA 94305, USA,\\
	\email{\{bruedigam, zacm\}@stanford.edu}
}

\maketitle              

\setcounter{equation}{39}
\section*{Rotational Gradient}
In \eqref{eqn:derivDiscRot} (Sec. \ref{sec:varrot}) we have used the \textit{rotational gradient} $\nabla^{\text{r}}$. We want to explain in more detail the idea of how to derive this gradient and why it is necessary. We start by revisiting the gradient of a function $f\colon \mathbb{R}^n \to \mathbb{R}$. The gradient of such a function is defined as
\begin{equation}\label{eqn:grad}
	\nabla_{\bs{x}}f(\bs{x}) = \begin{bmatrix}
		\derivb{f(\bs{x})}{x_1} \\ \derivb{f(\bs{x})}{x_2} \\ \vdots \\ \derivb{f(\bs{x})}{x_n}
	\end{bmatrix},
\end{equation}
and tells us what change of function value we get for changing values of $\bs{x}$. Stated more formally, we add a small deviation $\bs{\epsilon}$ to $\bs{x}$ and observe the relative change 
\begin{equation}
	\ddfrac{f(\bs{x}+\bs{\epsilon}) - f(\bs{x})}{\bs{\epsilon}}.
\end{equation}
Taking the limit of this relative change component-wise as $\bs{\epsilon}$ goes to zero yields
\begin{equation}\label{eqn:limit1}
	\lim_{\bs{\epsilon} \to \bs{0}} \ddfrac{f(\bs{x}+\bs{\epsilon}) - f(\bs{x})}{\bs{\epsilon}} = \nabla_{\bs{x}}f(\bs{x}),
\end{equation}
which is simply the gradient \eqref{eqn:grad}.

For rotations, a small deviation can be described as 
\begin{equation}
\bs{d} = \begin{bmatrix}
1 \\ \bs{\epsilon}
\end{bmatrix},
\end{equation}
since zero rotation would evaluate to the identity quaternion $[1 ~~ \bs{0}\hT]$. ``Adding'' this small deviation $\bs{d}$ to a quaternion $\bs{q}$ amounts to multiplication, similar to ``adding'' rotations with rotation matrices:
\begin{equation}\label{eqn:quatAdd}
\bs{q}\otimes\bs{d}.
\end{equation}
We can rewrite \eqref{eqn:quatAdd} as
\begin{equation}
\bs{q}\otimes\bs{d} = \begin{bmatrix}
q_w - \qno{v}\hT\bs{\epsilon}\\
\qno{v} + q_w\bs{\epsilon} + \qno{v}\times\bs{\epsilon}
\end{bmatrix}
= \bs{q} + \Lmat{\bs{q}}V\hT\bs{\epsilon}.
\end{equation}

Now, if we have a function $f(\bs{q})$ with some rotation described by a quaternion $\bs{q}$, we can again take a look at the change $f(\bs{q}\otimes\bs{d}) - f(\bs{q})$ for a small deviation $\bs{d}$. Taking the limit as in \eqref{eqn:limit1} and using the linear approximation 
\begin{equation}
f(\bs{q} + \Delta\bs{q}) = f(\bs{q}) + \Delta\bs{q}\hT\nabla_{\bs{q}}f(\bs{q}) ~~~~ (\text{true for small }\Delta\bs{q})
\end{equation}
we obtain the rotational gradient $\nabla^{\text{r}}_{\bs{q}}$ after simplification:
\begin{align}\label{eqn:limit2}
\lim_{\bs{\epsilon} \to \bs{0}} \ddfrac{f(\bs{q}\otimes\bs{d}) - f(\bs{q})}{\bs{\epsilon}} &= \lim_{\bs{\epsilon} \to \bs{0}} \ddfrac{f\left(\bs{q} + \Lmat{\bs{q}} V\hT\bs{\epsilon}\right) - f(\bs{q})}{\bs{\epsilon}}\\
&= \lim_{\bs{\epsilon} \to \bs{0}} \ddfrac{f(\bs{q}) + \bs{\epsilon}\hT V \LTmat{\bs{q}} \nabla_{\bs{q}}f(\bs{q}) - f(\bs{q})}{\bs{\epsilon}}\\
&= V \LTmat{\bs{q}} \nabla_{\bs{q}}f(\bs{q}) =: \nabla^{\text{r}}_{\bs{q}}f(\bs{q})
\end{align}
It is important to note that the rotational gradient $\nabla^{\text{r}}_{\bs{q}}f(\bs{q}) \in \mathbb{R}^3$ gives us information in the actual three-dimensional space of rotations despite a quaternion having four parameters. 

\section*{Rotational Equations of Motion}
In Sec. \ref{sec:varrot} we stated results for the rotational equations of motion derived from a variational principle. Here, we want to offer more detail on the derivation. The two main differences to the translational case are: (a) using an implicit unit norm constraint on quaternions to avoid tracking of the unit norm explicitly in a constraint and (b) applying the rotational gradient appropriately.

The quaternion angular velocity $\widetilde{\bs{\omega}}$ is defined as (cf. \eqref{eqn:angvel})
\begin{equation}\label{eqn:angvel2}
\widetilde{\bs{\omega}} = 2 \LTmat{\bs{q}} \dot{\bs{q}} = \begin{bmatrix}
	0\\\bs{\omega}
\end{bmatrix}.
\end{equation}
However, when using the discrete approximation of $\dot{\bs{q}}$ in \eqref{eqn:quatapprox}, we get a discrete angular velocity
\begin{equation}
\widetilde{\bs{\omega}}_k = \begin{bmatrix}
\widetilde{\omega}_{k,w}\\\bs{\omega}_{k}
\end{bmatrix},
\end{equation}
where $\widetilde{\omega}_{k,w}\neq 0$. 

We can update the orientation $\bs{q}_{k+1}$ at the next time step with this discrete angular velocity by discretizing and rearranging \eqref{eqn:angvel2}:
\begin{equation} 
\qno{k+1} = \ddfrac{\dt}{2}\Lmat{\qno{k}}\widetilde{\bs{\omega}}_k + \qno{k}.
\end{equation}
This update, however, will not maintain unit norm without an additional constraint because of $\widetilde{\omega}_{k,w}\neq 0$. We will therefore derive an expression for $\widetilde{\omega}_{k,w}$ similar to \cite{manchester_quaternion_2016}, that ensures unit norm:
\begin{align}
\left\lVert\qno{k+1}\right\rVert &= 1\\
\left\lVert\frac{\dt}{2}\Lmat{\qno{k}}\widetilde{\bs{\omega}}_k + \qno{k}\right\rVert &= 1\\
\Rightarrow\widetilde{\omega}_{k,w} &= \sqrt{\left(\tfrac{2}{\dt}\right)^2-\wno{k}\hT\wno{k}}-\ddfrac{2}{\dt}
\end{align}

This result was then used for the update rule in \eqref{eqn:updateOr} and to calculate the equations of motion in \eqref{eqn:dR}. 

Applying the rotational gradient to calculate the equations of motion is done as described in the previous section. However, there is a slight subtlety with the torque term in \eqref{eqn:discsumrot}. The correct derivative of this term is:
\begin{equation}
	\nabla^{\text{r}}_{\qno{k}}2\bs{\tau}_k\hT V \LTmat{\qno{k}} \qno{k} = 2 V \LTmat{\qno{k}} \Lmat{\qno{k}} V\hT \bs{\tau}_k
\end{equation}
Note that in this case, we are only taking the derivative with respect to the explicit $\bs{q}_k$ and we do not take the derivative of $\LTmat{\qno{k}}$. This special differentiation rule arises from the derivation of the generalized force corresponding to torque $\bs{\tau}$. More details can be found in \cite{manchester_quaternion_2016}.


\end{document}